\documentclass[a4paper,11pt]{article}
\pdfoutput=1

\usepackage[utf8]{inputenc}

\usepackage{amsthm,amsmath,amssymb}
\usepackage[numbers]{natbib}
\usepackage[colorlinks,citecolor=blue,urlcolor=blue]{hyperref}
\usepackage{enumerate}
\usepackage{algorithm}
\usepackage{algorithmic}
\usepackage{graphicx}
\usepackage{color}
\usepackage{rotating}
\usepackage{xspace}

\numberwithin{equation}{section}
\theoremstyle{plain}

\newtheorem{definition}{Definition}

\DeclareMathOperator{\prox}{prox}

\DeclareMathOperator{\diag}{diag}
\newcommand{\T}{\mathrm{T}}
\newcommand{\X}{\mathbf{X}}
\newcommand{\A}{\mathbf{A}}
\newcommand{\E}{\mathbf{E}}
\newcommand{\w}{\mathbf{w}}
\newcommand{\x}{\mathbf{x}}
\newcommand{\bt}{\mathbf{t}}

\newcommand{\y}{\mathbf{y}}
\newcommand{\C}{\mathbf{C}}
\newcommand{\I}{\mathbf{I}}
\newcommand{\M}[1]{\mathbf{#1}}
\renewcommand{\v}[1]{\mathbf{#1}}
\DeclareMathOperator*{\argmin}{arg\,min}
\DeclareMathOperator*{\argmax}{arg\,max}
\DeclareMathOperator*{\minimise}{minimise}
\renewcommand{\eqref}[1]{Equation~\ref{#1}}
\newcommand{\figref}[1]{Figure~\ref{#1}}
\newcommand{\algref}[1]{Algorithm~\ref{#1}}

\newcommand{\secref}[1]{Section~\ref{#1}}
\newcommand{\apxref}[1]{Appendix~\ref{#1}}
\def\proj{\mathrm{proj}}
\def\prox{\mathrm{prox}}
\newcommand{\Ie}{\textit{I.e.}\xspace}
\newcommand{\ie}{\textit{i.e.}\xspace}

\newcommand{\eg}{\textit{e.g.}\xspace}

\title{A general multiblock method for structured variable selection}
\author{Tommy L\"{o}fstedt, Fouad Hadj-Selem, Vincent Guillemot,\\
        Cathy Philippe, Nicolas Raymond, Edouard Duchesney,\\
        Vincent Frouin and Arthur Tenenhaus}

\begin{document}

\maketitle

\begin{abstract}
Regularised canonical correlation analysis was recently extended to more than
two sets of variables by the multiblock method Regularised generalised canonical
correlation analysis (RGCCA).

Further, Sparse GCCA (SGCCA) was proposed to address the issue of variable
selection. However, for technical reasons, the variable selection offered by
SGCCA was restricted to a covariance link between the blocks (\ie, with
\(\tau=1\)).

One of the main contributions of this paper is to go beyond the covariance link
and to propose an extension of SGCCA for the full RGCCA model
(\ie, with \(\tau\in[0, 1]\)). In addition, we propose an extension of SGCCA that
exploits structural relationships between variables within blocks.
Specifically, we propose an algorithm that allows structured and
sparsity-inducing penalties to be included in the RGCCA optimisation problem.

The proposed multiblock method is illustrated on a real three-block high-grade
glioma data set, where the aim is to predict the location of the brain tumours,
and on a simulated data set, where the aim is to illustrate the method's ability to
reconstruct the true underlying weight vectors.
\end{abstract}

\section{Introduction} \label{sec:introduction}

Regularised generalised canonical correlation analysis
(RGCCA)~\cite{Tenenhaus_and_Tenenhaus_2011} is a generalisation of
regularised canonical correlation analysis~\cite{Vinod_1976} to more
than two sets of variables. RGCCA relies on a sound theoretical foundation 
with a well-defined optimisation criterion, while at the same time allowing 
the analyst to incorporate prior knowledge or hypotheses about the relationships
between the blocks, as in PLS path modelling.

Sparse GCCA (SGCCA)~\cite{Tenenhaus_etal_2014} was recently proposed to address
the issue of variable selection. The RGCCA criterion was modified to include
\(\ell_1\) penalties on the outer weights vectors in order to promote sparsity.
For technical reasons concerning the RGCCA algorithm, the variable selection
offered by SGCCA was limited to the covariance link between blocks (\ie with all
\(\tau_k = 1\); see below for details). One of the main contributions of this paper
is to go beyond the covariance link and allow any \(\tau_k\in[0, 1]\).
More specifically, we present an extension of SGCCA that allows variable selection
to be performed for the full RGCCA model.

The sparsity induced by the \(\ell_1\) penalty does not take into account any
prior information on the relationships between variables within a block. Variables
could for instance belong to groups, or have spatial similarities (in \eg images),
and could therefore benefit from the ability to perform such \textit{structured
variable selection} instead. Further, different structured penalties could be added
to the different blocks, such that the regularisation adapts to the nature of the
blocks. For instance, naturally related groups of variables could be selected
together (or not at all), and noisy data could be constrained by spatial smoothness.
We will see examples of both of these kinds of penalties below.

Therefore, we propose in this work an extension of SGCCA that allows for the
exploitation of pre-given structural relationships between variables within blocks.
This is achieved by introducing structured complex penalties in the model. Such
penalties have recently become popular in machine learning and related
fields~\cite{Hadj-Selem_etal_2016_preprint} and encourage the resulting models to
have a particular structure. Structured complex penalties have previously been
considered in a two-block setting with canonical correlation
analysis~\cite{Chen_2011}. However, to combine such structured penalties with
RGCCA poses new challenges for the optimisation techniques used. In this paper we
propose a general multiblock algorithm that allows structured and sparsity-inducing
penalties to be included in the RGCCA model.

The authors presented the main ideas behind this work in~\cite{Lofstedt_etal_2016},
and we will here give the full theoretical exposition with all details of the
proposed method, including the derivation of a fast approach for projecting onto
the set induced by a quadratic penalty function, and an additional example on
simulated data.

\section{Method} \label{sect:method}

We consider several data matrices, \(\X_1, \ldots, \X_K\). Each \(n \times p_k\)
data matrix \(\X_k\) is called a block and represents a set of \(p_k\) variables
observed on \(n\) samples. The number of variables and the nature of the variables
usually differ from one block to another but the samples must be the same across
the blocks. We also associate to each matrix \(\X_k\) a column weight-vector \(\w_k\) 
of dimension \(p_k\).

Moreover, let \(\C=(c_{kj})\) be an adjacency matrix, where \(c_{kj}=1\) if the
blocks \(\X_k\) and \(\X_j\) are connected, and \(c_{kj}=0\) otherwise. The main
aim of RGCCA is to find block components, \(\y_k=\X_k\w_k\), for \(k=1, \ldots, K\),
that summarise the relevant information between and within the blocks,
while taking into account the structural connections between blocks defined by
the adjacency matrix. For that purpose, RGCCA is defined by the following
optimisation problem
\begin{align} \label{eq:loss_function_rgcca}
    \minimise_{\substack{\w_k\in\mathbb{R}^{p_k},\\k=1,\ldots,K}} \;\;& \varphi(\w_1,\ldots,\w_K) = -\sum_{k=1}^K\sum_{j=1}^K c_{kj}g(\text{Cov}(\X_k\w_k,\, \X_j\w_j)), \\
\label{eq:constraint_RGCCA}    \mathrm{subject~to} \;\;& \tau_k\|\w_k\|_2^2 + (1-\tau_k)\text{Var}(\X_k\w_k) = 1,
\end{align}
where the constraints are defined for all \(k=1, \ldots, K\); the function \(g\) is
called the inner-weighting scheme and can be any convex continuous function. Usually
the function \(g\) is one of the identity, \(g(x)=x\), called Horst's scheme, the
absolute value, \(g(x)=|x|\), called the Centroid scheme, or the square function,
\(g(x)=x^2\), called the Factorial scheme.

The regularisation parameters \(\tau_k\in[0,1]\) provide a way to control
the trade-off between maximising correlation and maximising covariance. The
above problem maximises (a function of) the covariance between connected
components if \(\tau_k = 1\), the correlation if \(\tau_k = 0\), and a
trade-off between covariance and correlation for all other values of
\(\tau_k\in(0, 1)\). The constraints defined by \eqref{eq:constraint_RGCCA}
can be expressed in matrix notation as \(\w_k^\T\M{M}_k\w_k\) where
\(\M{M}_k = \tau_k\I_{p_k} + \frac{1-\tau_k}{n-1}\X_k^\T\X_k\). We note that
\(\M{M}_k\) is positive-semidefinite, and positive-definite when \(\tau_k>0\) or
when \(\X_k\) is of full-rank.

The SGCCA framework limits the regularisation parameters to \(\tau_k = 1\) for
all \(k=1,\ldots,K\). This means that variable selection is only possible for
the special case with a covariance link in \eqref{eq:loss_function_rgcca}.
SGCCA is defined by the optimisation problem to
\begin{align} \label{eq:loss_function_l1}
    \minimise_{\substack{\w_k\in\mathbb{R}^{p_k},\\k=1,\ldots,K}} \;\;& \varphi(\w_1,\ldots,\w_K) \\
    \mathrm{subject~to}  \;\;& \|\w_k\|_2^2 = 1, \nonumber\\
    \phantom{subject~to} \;\;& \|\w_k\|_1 \leq s_k, \nonumber
\end{align}
where both constraints are defined for all blocks \(k=1, \ldots, K\);
the \(\|\w_k\|_1 = \sum_{j=1}^{p_k}|w_{k,j}|\) is the \(\ell_1\)-norm;
the \(s_k>0\) are the radii of the \(\ell_1\)-norm balls and determines the
amount of sparsity for \(\w_k\). The smaller \(s_k\) is, the larger the degree
of sparsity for \(\w_k\).

The \(\ell_1\) constraints are blind to any structure between the variables
within a block and are thus not able to account for \eg groups or similarities
between the variables in the RGCCA model.
We therefore propose to add \textit{structured penalties} to the objective
function. These structured penalties, account for such structured prior
knowledge or assumptions about the variables.

The optimisation problem that we consider is thus more general, and is defined by
\begin{align} \label{eq:loss_function_structure}
    \minimise_{\substack{\w_k\in\mathbb{R}^{p_k},\\k=1,\ldots,K}} \;\;& \varphi(\w_1,\ldots,\w_K) + \sum_{k=1}^K\omega_k\Omega_k(\w_k), \\
    \mathrm{subject~to} \;\; & \tau_k\|\w_k\|_2^2 + (1-\tau_k)\text{Var}(\X_k\w_k) \leq 1, \nonumber\\
\label{eq:constraintStructRGCCA} \phantom{subject~to} \;\;& \|\w_k\|_1 \leq s_k,
\end{align}
where both constraints are defined for all blocks \(k=1, \ldots, K\); the
functions \(\Omega_k\) are the structured penalties and the Lagrange
multipliers \(\omega_k\) are used as regularisation parameters.
The functions \(\Omega_k\) are convex, but not necessarily differentiable at
this point. This will be further discussed in \secref{sect:nesterov_smoothing}.

Unfortunately, since the objective function, $\varphi$, must be convex, we must
restrict the inner-weighting scheme, $g$, to Horst's scheme, \ie to the identity
$g(x)=x$.

Note that the equality in \eqref{eq:constraint_RGCCA} has been changed to an
inequality in \eqref{eq:constraintStructRGCCA}. The reason for this is that the
algorithm presented below requires the constraints to be convex. This is not
really a relaxation, however, since the Karush-Kuhn-Tucker conditions require
all constraints to be active at the solution, and it is always possible to find
constraint parameters, \(s_k\), such that both constraints are active for each
block~\cite{Witten_etal_2009}.

When the structured penalties, \(\Omega_k\), are convex,
\eqref{eq:loss_function_structure} is a multiconvex function with convex
constraints. This means that the function is convex with respect to one block 
weight vector \(\w_k\) at the time. \Ie if we consider
\(\w_1,\ldots,\w_{k-1},\w_{k+1},\ldots,\w_K\) constant, the function is convex
with respect to \(\w_k\).

However, the structured penalties are usually neither smooth nor separable, \ie
they can not be written as a separable sum. We can therefore not minimise the
penalties together with the smooth loss function by using smooth minimisation
algorithms, and have to revert to non-smooth minimisation algorithms
such as \eg proximal methods. However, to compute the proximal operator of the
structured penalty, we rely on separability to minimise the
proximal definition coordinate-wise. Without separability, the system does not
usually have an explicit solution and is therefore difficult to solve.

This means that it may be very difficult to find a minimum in the general case.
Solutions exist for some particular structured penalties, but they are tailored
towards a particular formulation, and can not be used for the general problem
that was defined in \eqref{eq:loss_function_structure}. We therefore adapt a
very efficient smoothing technique proposed by Nesterov~\cite{Nesterov_2004} to
resolve both the non-smoothness and non-separability issues for a very wide and
general class of structured penalties. This smoothing technique is presented in
the next section.


\subsection{Nesterov's smoothing technique} \label{sect:nesterov_smoothing}

The structured penalties, \(\Omega_k\), considered in this paper are convex but
possibly non-differentiable. The functions \(\Omega_k\) must fit the framework
of Nesterov, as described in~\cite{Hadj-Selem_etal_2016_preprint}, and will be
written in the form
\[
    \Omega_k(\w_k) = \sum_{G=1}^{\mathcal{G}_k} \|\A_{k,G}\w_k\|_q
                   = \sum_{G=1}^{\mathcal{G}_k} \max_{\|\boldsymbol{\alpha}_{k,G}\|_{q'} \leq 1} \langle\boldsymbol{\alpha}_{k,G}\,|\,\A_{k,G}\w_k\rangle,
\]
in which \(\mathcal{G}_k\) is the number of \textit{groups} for the particular
function \(\Omega_k\). A \textit{group} constitutes the variables with
associated non-zero entries in \(\A_{k,G}\) (this would be \eg the pixels or
voxels associated with the gradient at a particular point in total variation,
or a group of related variables in group \(\ell_{1,q}\)). The \textit{group
matrix} \(\A_{k,G}\) is a linear operator for group \(G\) associated with the
function \(\Omega_k\), for \(k=1,\ldots,K\). The function \(\|\cdot\|_q\) is the
standard \(\ell_q\)-norm defined on \(\mathbb{R}^p\) by
\(\|\x\|_q = \sqrt[q]{\sum_{j=1}^p |x_j|^q}\), with the
associated dual norm \(\|\cdot\|_{q'}\) for \(q, q' \geq 1\). Nesterov's
smoothing technique~\cite{Hadj-Selem_etal_2016_preprint, Nesterov_2004} is formally
defined as follows.
\begin{definition} \label{def:nesterov}
    Let \(\Omega\) be a convex function. A sufficient condition for the
    application of Nesterov's smoothing technique is that \(\Omega\) can be
    written in the form
    \begin{equation*} 
        \Omega(\w) = \max_{\boldsymbol{\alpha} \in K} \, \langle\boldsymbol{\alpha}\,|\,\A\w\rangle,
    \end{equation*}
    for all \(\w \in \mathbb{R}^p\), with \(K\) a compact convex set in a
    finite-dimensional vector space and \(\A\) a linear operator between two
    finite-dimensional vector spaces. Given this expression for \(\Omega\),
    Nesterov's smoothing is defined as
    \begin{equation*} 
        \widehat{\Omega}(\mu, \w) = \langle\boldsymbol{\alpha}^*\,|\,\A\w\rangle - \frac{\mu}{2}\|\boldsymbol{\alpha}^*\|_2^2,
    \end{equation*}
    for all \(\w \in \mathbb{R}^p\), with \(\mu\) a positive real smoothing
    parameter and where
    \begin{equation*}
        \boldsymbol{\alpha}^* = \argmax_{\boldsymbol{\alpha} \in K}\left\{\langle\boldsymbol{\alpha}\;|\,\A\w\rangle - \frac{\mu}{2}\|\boldsymbol{\alpha}\|_2 ^2\right\}.
    \end{equation*}
\end{definition}

When smoothing the functions \(\widehat{\Omega}_k\) this way, we obtain
\begin{equation*}
    \lim_{\mu_k\rightarrow0}\widehat{\Omega}_k(\mu_k, \w_k) = \Omega_k(\w_k).
\end{equation*}
An immediate consequence is that since the functions \(\widehat{\Omega}_k\) are
convex and differentiable they may, for a sufficiently small value of \(\mu_k\),
be used instead of \(\Omega_k\).

The gradients of the Nesterov smoothed functions,
\(\widehat{\Omega}_k(\mu_k, \w_k)\), with respect to the corresponding \(\w_k\)
are
\[
    \nabla_{\w_k}\widehat{\Omega}_k(\mu, \w_k) = \A_k^\T\boldsymbol{\alpha}_k^*.
\]
The gradients are Lipschitz continuous with Lipschitz constant
\[
    L\big(\nabla_{\w_k}\widehat{\Omega}_k(\mu, \w_k)\big) = \frac{\|\A_k\|_2^2}{\mu},
\]
where \(\|\A_k\|_2\) is the spectral norm of \(\A_k\).

\subsection{Reformulation of the objective} \label{sect:reformulated_objective}

Nesterov's smoothing technique allow us to have a smooth objective function with
convex constraints. In order to find a minimiser to
\eqref{eq:loss_function_structure} we must first alter the formulation slightly.

The constraints are rephrased as follows: We construct the sets
$\mathcal{P}_k$ and $\mathcal{S}_k$ as
$\mathcal{P}_k = \{\x\in\mathbb{R}^{p_k} \;|\; \|\x\|_1 \leq s_k\}$ and
$\mathcal{S}_k = \{\x\in\mathbb{R}^{p_k} \;|\;  \w_k^\T\M{M}_k\w_k \leq 1\}$,
and then form the intersection set,
\(\mathcal{W}_k = \left\{\x \;|\; \x \in \mathcal{P}_k \cap \mathcal{S}_k \right\}\).
Therefore, we are interested in block weight vectors \(\w_k\) such that
\(\w_k \in \mathcal{W}_k\).
We will suppose that \(\mathcal{W}_k \neq \emptyset\) and in fact assume that at
least \(\v{0} \in \mathcal{W}_k\). We note that all \(\mathcal{W}_k\) are convex
sets, since \(\mathcal{P}_k\) and \(\mathcal{S}_k\) are convex sets.

The optimisation problem in \eqref{eq:loss_function_structure}, and the final
optimisation problem that we will consider in this paper, can thus be stated as
\begin{align} \label{eq:loss_function_reformulated}
    \minimise_{\w_1, \ldots, \w_K}~ & \widehat{f}(\w_1,\ldots,\w_K) = \varphi(\w_1,\ldots,\w_K) + \sum_{k=1}^K\omega_k\widehat{\Omega}_k(\mu_k, \w_k) \\
    \mathrm{subject~to~~} & \w_k \in \mathcal{W}_k, \mathrm{~~} k=1,\ldots,K. \nonumber
\end{align}
This single indicative constraint for each block is equivalent to the
constraints in \eqref{eq:loss_function_structure}. When \(\mu_k \rightarrow 0\),
the two problems in \eqref{eq:loss_function_structure} and
\eqref{eq:loss_function_reformulated} are equivalent.

The partial gradients of the objective function in
\eqref{eq:loss_function_reformulated} with respect to each \(\w_k\) are
\begin{align} \label{eq:gradient_of_objective}
    \nabla_{\w_k}&\widehat{f}(\w_1,\ldots,\w_K) = \\
                            & -\sum_{j=1}^K c_{kj}g'(\text{Cov}(\X_k\w_k,\, \X_j\w_j))\frac{1}{n-1}\X_k^\T\X_j\w_j + \omega_k\A_k^\T\boldsymbol{\alpha}_k^*, \nonumber
\end{align}
for all \(k=1,\ldots,K\), where we let
\(\text{Cov}(\X_k\w_k,\, \X_j\w_j) = \frac{1}{n-1}\w_k^\T\X_k^\T\X_j\w_j\), \ie
the unbiased sample covariance.

\section{Algorithm} \label{sec:algorithm}

In order to find a solution to the problem in
\eqref{eq:loss_function_reformulated}, a multi-convex function with an
indicative constraint over a convex set, we minimise it over several
different parameter vectors (\ie \(\w_1, \ldots, \w_K\)) by updating each of
the parameter vectors in turn while keeping the others fixed. \Ie we minimise
the function over one parameter vector at the time and treat the other parameter
vectors as constants during this minimisation. If each update improves the
function value, gradually the function will be (locally) optimised over the
entire set of parameter vectors. This principle is called block
relaxation~\cite{DeLeeuw_1994}. The algorithm we present in
\algref{alg:RGCCA_algorithm} is related to the algorithm presented
in~\cite{Witten_etal_2009}. However, several details need to be introduced
before we discuss the proposed algorithm further.


\subsection{Projection operators}

We see in \algref{alg:RGCCA_algorithm} that orthogonal projections onto the
convex sets \(\mathcal{W}_k\) are required at each iteration of the algorithm,
and for each block.

The projection onto the intersection of two convex sets,
\(\mathcal{W}=\mathcal{P}\cap\mathcal{S}\), is formulated as the unique point
that minimises the problem
\begin{align} \label{eq:proj_loss_function}
    \proj_{\mathcal{W}_k}(\x) &= \argmin_{\y \in \mathcal{W}_k} \frac{1}{2}\|\y - \x\|_2^2 \\
        &= \argmin_{\y \in \mathbb{R}^{p_k}} \frac{1}{2}\|\y - \x\|_2^2 + \iota_{\mathcal{W}_{k}}(\y). \nonumber
\end{align}
where \(\iota_{\mathcal{W}_{k}}\) is the indicator function over
\(\mathcal{W}_{k}\), \ie
\begin{equation*}
    \iota_{\mathcal{W}_{k}}(\x) = \begin{cases}
                   0      & \text{if~} \x \in \mathcal{W}_{k}, \\
                   \infty & \text{otherwise}.
               \end{cases}
\end{equation*}
We note that the right-most side of \eqref{eq:proj_loss_function} is in fact
the proximal operator of the indicator function over the set \(\mathcal{W}_k\),
and thus that the proximal operator of the indicator function is the projection
onto the corresponding set.

The projection onto the intersection \(\mathcal{P}\cap\mathcal{S}\) can be
computed using Dykstra's projection algorithm~\cite{Combettes_Pesquet_2011},
stated in \algref{alg:Dykstra_projection}. The sequence
\((\x^{(s)})_{s \in \mathbb{N}}\) generated by \algref{alg:Dykstra_projection}
converges to the unique point that is the solution to
\eqref{eq:proj_loss_function}.

We thus use \algref{alg:Dykstra_projection} to find the projection onto the
intersection of the two sets \(\mathcal{P}\) and \(\mathcal{S}\). This is
necessary in order to enforce the constraint in
\eqref{eq:loss_function_reformulated}. Three key points need to be explained in
order to make \algref{alg:Dykstra_projection} clear:
\begin{enumerate}[(i)]
    \item the projection onto \(\mathcal{P}\) (Line 3),
    \item the projection onto \(\mathcal{S}\) (Line 5), and
    \item the stopping criterion (Line 7).
\end{enumerate}
These points are discussed in the following subsections.

\begin{algorithm}
    \caption{Dykstra's projection algorithm}
    \label{alg:Dykstra_projection}
    \begin{algorithmic}[1]
        \REQUIRE \(\x^{(0)}\), \(\mathcal{P}\), \(\mathcal{S}\), \(\varepsilon>0\)
        \ENSURE \(\x^{(s)}\in\mathcal{P}\cap\mathcal{S}\)
        \STATE \(\v{p}^{(0)} \leftarrow 0\),\; \(\v{q}^{(0)} \leftarrow 0\)
        \FOR{\(s=0,1,2,\ldots\)}
            \STATE \(\v{y}^{(s)} = \proj_{\mathcal{P}}(\v{x}^{(s)} + \v{p}^{(s)})\)
            \STATE \(\v{p}^{(s+1)} = \x^{(s)} + \v{p}^{(s)} - \v{y}^{(s)}\)
            \STATE \(\x^{(s+1)} = \proj_{\mathcal{S}}(\v{y}^{(s)} + \v{q}^{(s)})\)
            \STATE \(\v{q}^{(s+1)} = \v{y}^{(s)} + \v{q}^{(s)} - \x^{(s+1)}\)
            \IF{\(\max\big(\|\x^{(s+1)} - \proj_{\mathcal{P}}(\x^{(s+1)})\|_2,\;
                           \|\x^{(s+1)} - \proj_{\mathcal{S}}(\x^{(s+1)})\|_2\big) \leq \varepsilon\)}
                \STATE \textbf{break}
            \ENDIF
        \ENDFOR
    \end{algorithmic}
\end{algorithm}

\subsubsection{Projection onto $\mathcal{P}$}

The projection onto the \(\ell_1\) ball is achieved by utilising a very
efficient method presented in \eg~\cite{van_den_Berg_etal_2008}. This method
uses the proximal operator of \(\|\cdot\|_1\), the \emph{soft thresholding
operator}~\cite{Parikh_Boyd_2013}, which is defined as
\begin{equation} \label{eq:soft_thresholding}
    \left(\prox_{\lambda_k\|\cdot\|_1}(\x)\right)_i =
    \begin{cases}
        x_i - \lambda_k, & \mbox{if } x_i > \lambda_k, \\
        0, & \mbox{if } |x_i| \leq \lambda_k, \\
        x_i - \lambda_k, & \mbox{if } x_i < -\lambda_k.
    \end{cases}
\end{equation}

This leads to the problem of finding a solution, \(\lambda_k^*\), to the
equation
\begin{equation} \label{eq:l1_projection_problem}
    \sum_{i=1}^{p_k}(|x_i| - \lambda_k^*)_+ = s_k,
\end{equation}
where \((x)_+ = \max(0,\,x)\).

Using the parameter \(\lambda_k^*\) with the proximal operator results in the
projection onto an \(\ell_1\) ball of radius \(s_k\). \Ie
\begin{equation*}
    \proj_{\mathcal{P}_k}(\x) = \prox_{\lambda_k^*\|\cdot\|_1}(\x),
\end{equation*}
where thus \(\lambda_k^*\) is the solution of \eqref{eq:l1_projection_problem}.

The method we use makes the observation that if the absolute values of \(\x\)
are sorted, the solution to \eqref{eq:l1_projection_problem} is found between
two consecutive values of the sorted absolute \(x_i\). The exact optimal value
is then found by simply interpolating linearly (because of the nature of the
Lagrange dual function of \eqref{eq:soft_thresholding}) between those two
values. See~\cite{van_den_Berg_etal_2008} for the details.

\subsubsection{Projection onto $\mathcal{S}$}

The \textit{$\mathcal{S}$ constraint} is quadratic, which means its proximal
operator is
\begin{align} \label{eq:RGCCA_prox}
    \prox_{\lambda_k}(\x) &= \argmin_{\y\in\mathbb{R}^{p_k}} \frac{1}{2}\|\y - \x\|_2^2 + \lambda_k \y^\T\M{M}_k\y \\
                          &= \left(\I_{p_k} + 2\lambda_k\M{M}_k\right)^{-1}\x. \nonumber
\end{align}
Let \(\lambda_k^*\) be the smallest \(\lambda_k\) such that \(\y^\T\M{M}_k\y \leq 1\), then
\[
    \proj_{\mathcal{S}_k}(\x) = \prox_{\lambda_k^*}(\x).
\]
It is not feasible to compute this projection by using \eqref{eq:RGCCA_prox}
directly. Especially not when the number of variables is very large. \Ie it is
not feasible to numerically find this \(\lambda_k^*\) directly from
\eqref{eq:RGCCA_prox} because of the computational effort required by the
inverse. We have therefore instead devised a very efficient algorithm that
rephrases the problem and then utilises the Newton-Raphson method to compute
\(\lambda_k^*\) from a simple univariate auxiliary function that only depends on
the eigenvalues of \(\M{M}_k\). See~\apxref{apx:RGCCA_constraint} for the
details.

\subsubsection{Stopping Criterion}

Since the projection on Line~\ref{alg:RGCCA_algorithm:line:proj} of
\algref{alg:RGCCA_algorithm} is approximated (using
\algref{alg:Dykstra_projection}), we are actually performing an inexact
projected gradient descent~\cite{Schmidt_etal_2011}. We must therefore make
sure that the approximation is close enough that we still converge to the
minimum of the objective function.

At step \(s\) of \algref{alg:RGCCA_algorithm}, after projection onto
\(\mathcal{W}\) with \algref{alg:Dykstra_projection}, the following inequality
must be respected in order to ensure convergence to the minimum of the objective
function:
\[
    \|\w^{(s+1)} - \proj_{\mathcal{W}}\big(\w^{(s+1)}\big)\|_2 < \varepsilon^{(s)},
\]
where the precision, \(\varepsilon^{(s)}\), must decrease like
\(\mathcal{O}\big(1/i_k^{4+\delta}\big)\), for any \(\delta>0\), and where
\(i_k\) is the iteration counter of FISTA for block \(k\). This follows from
Proposition~2 in~\cite{Schmidt_etal_2011} (for FISTA, and Proposition~1 for
ISTA).

Since we can not compute the distance
\(\|\w^{(s+1)} - \proj_{\mathcal{W}}\big(\w^{(s+1)}\big)\|_2\) directly (this
requires a solution to the main problem we are trying to solve) and since
\(\mathcal{W}\) is the intersection of the convex sets \(\mathcal{P}\) and
\(\mathcal{S}\), we may approximate it by
\[
    \max\Big(\|\x^{(s+1)} - \proj_\mathcal{P}(\x^{(s+1)})\|_2,\; \|\x^{(s+1)} - \proj_\mathcal{S}(\x^{(s+1)})\|_2\Big),
\]
because of the well-known relation that
\begin{align}
    \|&\x^{(s+1)} - \proj_\mathcal{W}(\x^{(s+1)})\|_2 \\
                 &\;< \kappa \cdot \max\Big(\|\x^{(s+1)} - \proj_\mathcal{S}(\x^{(s+1)})\|_2,\; \|\x^{(s+1)} - \proj_\mathcal{P}(\x^{(s+1)})\|_2\Big), \nonumber
\end{align}
for some positive real scalar \(\kappa\).

\subsection{Algorithm for Structured Variable Selection in RGCCA}

We are now ready to discuss the full multiblock accelerated projected gradient
method, a generalised RGCCA minimisation algorithm. This algorithm is presented
in \algref{alg:RGCCA_algorithm}.

\begin{algorithm}[!h]
    \caption{Algorithm for structured variable selection in RGCCA}
    \label{alg:RGCCA_algorithm}
    \begin{algorithmic}[1]
        \REQUIRE \(\widehat{f}\),\, \(\nabla\widehat{f}\),\, \(\v{w}_k=\v{w}_k^{(0)}\in\mathcal{W}_k\),\, \(\varepsilon>0\)
        \ENSURE \(\v{w}_k^{(s)}\in\mathcal{W}_k\) such that \(\varepsilon\in\partial\widehat{f}(\v{w}_1^{(s)},\ldots,\v{w}_K^{(s)})\)
        \REPEAT
            \FOR{\(k=1\) \TO \(K\)}
                \STATE \(\v{w}_k^{(1)} = \v{w}_k^{(0)} = \v{w}_k\)
                \FOR{\(s=1,2,\ldots\)}
                    \STATE $\v{y} = \v{w}_k^{(s)} + \frac{k-2}{k+1}\big(\v{w}_k^{(s)} - \v{w}_k^{(s-1)}\big)$
                    \STATE \(\v{w}_k^{(s+1)} = \mathrm{proj}_{\mathcal{W}_k}\Big(\v{y} - t_k\nabla_{\v{w}_k^{(s)}}\widehat{f}\big(\v{w}_1^{(s)},\ldots,\v{y},\ldots,\v{w}_K^{(s)}\big)\Big)\) \label{alg:RGCCA_algorithm:line:proj}
                    \IF{\(\|\v{w}_k^{(s+1)} - \v{y}\|_2 \leq t_k\varepsilon\)}
                        \STATE \textbf{break}
                    \ENDIF
                \ENDFOR
                \STATE \(\v{w}_k = \v{w}_k^{(s+1)}\)
            \ENDFOR
        \UNTIL{\(\big\|\v{w}_k - \mathrm{proj}_{\mathcal{W}_k}\big(\v{w}_k - t_k\nabla_{\v{w}_k}\widehat{f}(\v{w}_1,\ldots,\v{w}_K)\big)\big\|_2 < t_k\varepsilon\), \textbf{for all} \(k=1,\ldots,K\)} \label{alg:RGCCA_algorithm:line:main_stopping_criterion}
    \end{algorithmic}
\end{algorithm}

Any appropriate minimisation algorithm can be used in the inner-most loop of
\algref{alg:RGCCA_algorithm}. We use the fast iterative shrinkage-thresholding
algorithm (FISTA)~\cite{Beck_Teboulle_2009, Beck_Teboulle_2009b}, since it has
the optimal (for first-order methods) convergence rate of \(\mathcal{O}(1/s^2)\),
where \(s\) is the iteration count.

FISTA requires a step size, \(t_k\), for each block and each step of the
iterative algorithm, as seen on Line~\ref{alg:RGCCA_algorithm:line:proj} of
\algref{alg:RGCCA_algorithm}. If all partial gradients of the objective function
in \eqref{eq:loss_function_reformulated}, \ie the gradients in
\eqref{eq:gradient_of_objective}, are Lipschitz continuous, then we can compute
the step size directly. In that case, the step sizes, \(t_k\), are computed as
the reciprocal of the sum of the Lipschitz constants of the gradients, as
explained in~\cite{Hadj-Selem_etal_2016_preprint}. \Ie, such that
\begin{align}
    t_k &= \Big(L\big(\nabla_{\w_k}\widehat{f}\big)\Big)^{-1} \nonumber\\
        &= \Big(L\big(\nabla_{\w_k}\phi\big) + L\big(\nabla_{\w_k}\omega_k\widehat{\Omega}_k(\mu, \w_k)\big)\Big)^{-1} \nonumber\\
        &= \Bigg(L\big(\nabla_{\w_k}\phi\big) + L\bigg(\frac{\omega_k\|\A_k\|_2^2}{\mu_k}\bigg)\Bigg)^{-1}, \nonumber
\end{align}
where \(\mu_k\) is the parameter for the Nesterov smoothing; the partial gradients
are from the the loss function in \eqref{eq:loss_function_reformulated}, \ie the
Lipschitz constants of the partial gradients in
\eqref{eq:gradient_of_objective}. If some gradient is not Lipschitz continuous,
or if the sum of Lipschitz constants would be zero, the step size can also be
found efficiently by using backtracking line search.

Note that the main stopping criterion on
Line~\ref{alg:RGCCA_algorithm:line:main_stopping_criterion} is actually
performing a step of the iterative soft-thresholding algorithm (ISTA).
This stopping criterion can easily be explained as follows: Assume a function on
the form
\[
    f(\x) = g(\x) + h(\x),
\]
where \(g\) is smooth and convex, and \(h\) is convex, but non-smooth, and whose
proximal operator is known. We recall and rewrite the ISTA descent step,
\begin{align}
    \x^{(s+1)} &= \prox_{th}\big(\x^{(s)} - t\nabla g(\x^{(s)})\big) \nonumber\\
               &= \x^{(s)} - tG_t(\x^{(s)}), \nonumber
\end{align}
where
\[
    G_t(\x) = \frac{1}{t}\Big(\x - \prox_{th}\big(\x - t\nabla g(\x)\big)\Big).
\]
Thus, it is clear that convergence has been achieved if \(G_t(\x^{(s)})\) is
\textit{small}. In fact, it follows from the definition of subgradients and the
optimality condition of proximal operators~\cite{Qin_etal_2013} that
\[
    G_t(\x) \in \nabla g(\x) + \partial h\big(\x - tG_t(\x)\big),
\]
and that \(G_t(\x) = \v{0}\) if and only if \(\x\) minimises
\(f(\x) = g(\x) + h(\x)\).

\section{Examples}

We will illustrate the proposed method by two examples. The first example is on
a real three-block glioma data set where the aim is to predict the location of
brain tumours from gene expression (GE) and comparative genomic hybridisation (CGH)
data. The second example is on a simulated data set where the aim is to see if it
is possible to reconstruct the true underlying weights.

\subsection{Glioma data set}

We illustrate the proposed method by predicting the location of brain tumours
from GE and CGH data~\cite{Philippe_2012}. The problem is one with three blocks:
GE (\(\X_1 \in \mathbb{R}^{53\times 15702}\)),
CGH (\(\X_2 \in \mathbb{R}^{53\times 41996}\)) and a dummy matrix encoding the
locations (\(\X_3 \in \mathbb{R}^{53\times 3}\)). The locations were: The brain
stem (DIPG), central nuclei (Midline) and supratentorial (Hemisphere).

The purpose of this example is to show the versatility of the proposed method,
and to show how it can be used to build an RGCCA model with both complex
penalties and sparsity-inducing constraints, and analyse data related to the
data used in~\cite{Tenenhaus_etal_2014}.

\begin{figure}[!b]
    \begin{center}
        \includegraphics[width=1.0\textwidth]{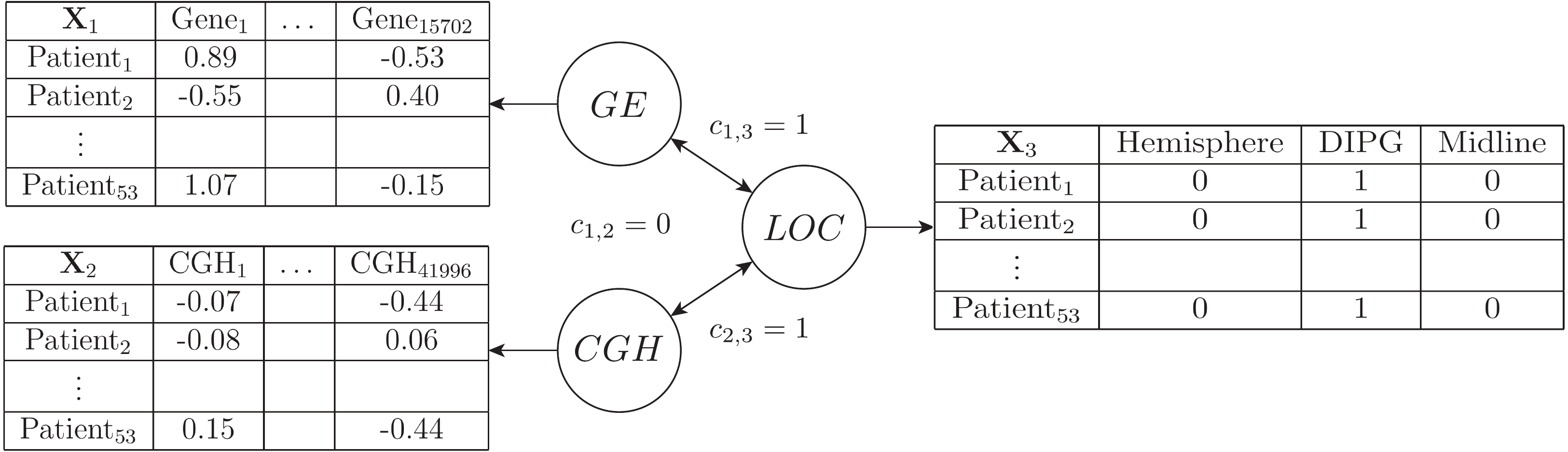}
        \caption{Path diagram for the prediction model. \(\X_1\) (GE) is connected to \(\X_3\) (location), \(\X_2\) is also connected to \(\X_3\) but \(\X_1\) and \(\X_2\) are not connected.}
        \label{fig:model_design}
    \end{center}
\end{figure}

The relation design was, the between-block connections, were chosen to be oriented
towards prediction, and is illustrated in \figref{fig:model_design}.
Therefore,  \(\X_1\) and \(\X_2\) are connected to \(\X_3\)
(\(c_{13} = c_{23} = 1)\), but there is no connection between \(\X_1\) and
\(\X_2\) (\ie, \(c_{12} = 0\)). This design tends to focus on models where
prediction can be made even if there is no relation between the predictor blocks,
and in~\cite{Tenenhaus_etal_2014} it was indeed the case on equivalent data that
this design yielded the best prediction rates among similar designs.

An \(\ell_1\) and a group \(\ell_{1,2}\)~\cite{Silver_and_Giovanni_2012,
Qin_etal_2013} constraint were added to the GE block, \(\X_1\). An \(\ell_1\)
together with a total variation~\cite{Michel_etal_2011}, constraint were added
to the CGH, \(\X_2\), block in order to smooth the often noisy CGH data.


The regularisation constants, \(\tau_k\), for \(k=1,2,3\), were computed using
the method of Sch\"{a}fer and Strimmer~\cite{Schafer_and_Strimmer_2005}, and were
\(\tau_1=1.0\) and \(\tau_3=1.0\), for blocks \(\X_1\) and \(\X_3\),
respectively. For \(\X_2\) we were unable to compute the regularisation
constant, because of the large size of the data. We therefore instead used the mean
of ten regularisation constants, computed from random samples of $41\,996/2=20\,998$
variables each, and rounded to one decimal point. The computed mean was
\(\tau_2=0.2623\ldots \approx 0.3\).

The other constants were found by grid search with 7-fold cross-validation. The
cross-validation procedure maximised the statistic
\begin{align}
    \mathrm{R^2_{pred,\X_3}}
      &= \mathrm{R^2_{pred,\X_1\rightarrow\X_3}} \cdot \mathrm{R^2_{pred,\X_2\rightarrow\X_3}} \\
      &= \left(1 - \frac{\|\widehat{\X}_{1\rightarrow 3} - \X_3\|^2_F}{\|\X_3\|^2_F}\right)
       \cdot
         \left(1 - \frac{\|\widehat{\X}_{2\rightarrow 3} - \X_3\|^2_F}{\|\X_3\|^2_F}\right) \nonumber
\end{align}
in order to force a high prediction rate from both $\X_1$ and $\X_2$.
\Ie, \(\mathrm{R^2_{pred,\X_3}}\) is the combined prediction rate from
the models of \(\X_1\) and \(\X_2\), where the product forces both blocks to
predict \(\X_3\) well. By
\(\widehat{\X}_{1\rightarrow 3}\) we denote the prediction of the locations,
encoded in the dummy matrix \(\X_3\), from the model of \(\X_1\) and by
\(\widehat{\X}_{2\rightarrow 3}\) we denote the prediction of \(\X_3\) from the
model of \(\X_2\). The \(\|\cdot\|^2_F\) denotes the squared Frobenius norm.

The predictions were computed in the traditional way using the inner (or
structural) relation~\cite{Tenenhaus_etal_2005, Sanchez_2013, Wegelin_2000},
\ie that the assumed relation between corresponding latent variables is linear,
\[
    \widehat{\v{t}}_{k \rightarrow j} = \v{t}_k b_{k \rightarrow j} + \boldsymbol{\varepsilon}_{k \rightarrow j},
\]
where
\[
    b_{k \rightarrow j} = \frac{\v{t}_k^\T\v{t}_j^{\vphantom{T}}}{\v{t}_k^\T\v{t}_k^{\vphantom{T}}},
\]
is a regression coefficient. Multiple regression is performed
when there are several latent variables.

Finally, we predict with
\[
    \widehat{\X}_{k\rightarrow 3}^{\vphantom{T}} = \M{T}_k^{\vphantom{T}}(\M{T}_k^\T\M{T}_k^{\vphantom{T}})^{-1}\M{T}_k^\T\M{T}_3^{\vphantom{T}}\M{W}_3^\T,
    = \widehat{\M{T}}_{k\rightarrow 3}^{\vphantom{T}}\M{W}_3^\T
\]
where \(\M{T}_k = [\X_k\v{w}_{k,1}|\X_k\v{w}_{k,2}|\cdots|\X_k\v{w}_{k,A}]\) are the \(A\)
latent variables for block \(k\) and \(\M{W}_3 = [\v{w}_{3,1}|\cdots|\v{w}_{3,A}]\)
are the \(A\) weight vectors for block 3.

The criterion of the final model, as found by the grid search, was
\[
    \mathrm{R^2_{pred,\X_3}} =
        \mathrm{R^2_{pred,\X_1\rightarrow\X_3}} \cdot \mathrm{R^2_{pred,\X_2\rightarrow\X_3}} =
        0.51 \cdot 0.47 \approx 0.24.
\]
These numbers were computed as the means of 7-fold cross-validation.

The regularisation constant for the group \(\ell_{1,2}\) penalty was thus deemed by
the grid search to be \(\omega_1=0.35\) and the \(\ell_1\) norm constraint had a
radius of \(s_1=13\). The regularisation constant for total variation was
\(\omega_2=0.004\), and the \(\ell_1\) norm constraint had a radius of
\(s_2=10.1\).

The optimisation problem that we considered in this example was thus
\begin{align}
    \minimise_{\substack{\w_k\in\mathbb{R}^{p_k},\\k=1,2,3}} \;& \widehat{f}(\v{w}_1,\v{w}_2,\v{w}_3) = -\text{Cov}(\X_1\v{w}_1, \X_3\v{w}_3) -\text{Cov}(\X_2\v{w}_2, \X_3\v{w}_3) \nonumber\\[-1em]
                                      &\qquad\qquad\qquad\quad\; + 0.35\cdot\widehat{\Omega}_{GL}(\mu_1, \v{w}_1) + 0.004\cdot\widehat{\Omega}_{TV}(\mu_2, \v{w}_2), \nonumber\\
    \mathrm{subject~to} \;& \w_1 \in \{\x \in \mathbb{R}^{p_1} \;|\; \|\x\|_1 \leq 13 \land \|\x\|_2^2 \leq 1\}, \nonumber\\
    \phantom{subject~to} \;& \w_2 \in \{\x \in \mathbb{R}^{p_2} \;|\; \|\x\|_1 \leq 10.1 \land 0.3\cdot\|\x\|_2^2 + 0.7\cdot\text{Var}(\X_2\x) \leq 1\}, \nonumber\\
    \phantom{subject~to~} & \w_3 \in \{\x \in \mathbb{R}^{p_3} \;|\; \|\x\|_2^2 \leq 1\}, \nonumber \\
\end{align}
in which \(\mu_1=\mu_2=5\cdot10^{-4}\). Two components were extracted using the
deflation scheme~\cite{Tenenhaus_etal_2014}
\[
    \X_k \leftarrow \X_k - \frac{\X_k^{\vphantom{T}}\w_k^{\vphantom{T}}\w_k^\T}{\w_k^\T\w_k^{\vphantom{T}}}.
\]

\subsection{Simulated data}

In order to illustrate one of the main benefits of the proposed method, we
performed a simulation study in which we compared the differences between the
weight vectors of the proposed method and ``regular'' unpenalised RGCCA to the
weight vectors used when generating the simulated data.

We generated data for two blocks, \(\X_1\in\mathbb{R}^{50\times150}\) and
\(\X_2\in\mathbb{R}^{50\times100}\), defined by the models
\[
    \X_1 = \bt_1\w_1^\T + \E_1,
\]
\[
    \X_2 = \bt_2\w_2^\T + \E_2,
\]
where \(\bt_1\sim\mathcal{N}(\mathbf{0}, \I_{50})\),
\(\bt_2 \sim \mathcal{N}(\bt_1, 0.01^2\I_{50})\); the columns of \(\E_1\),
were random normal, \(\E_{1,j_1}\sim\mathcal{N}(\mathbf{0}, 0.15^2\I_{50})\), and
similarly for the columns of \(\E_2\), such that
\(\E_{2,j_2}\sim\mathcal{N}(\mathbf{0}, 0.2^2\I_{50})\).

We then built an RGCCA model using \eqref{eq:loss_function_rgcca} and a
regularised RGCCA model using \eqref{eq:loss_function_reformulated}. The
regularised RGCCA model had a total variation penalty and an \(\ell_1\) constraint
for \(\X_1\), and a group \(\ell_{1,2}\) penalty for \(\X_2\). The block \(\X_2\)
was not given an \(\ell_1\) constraint in order to illustrate the variable
selection that the group \(\ell_{1,2}\) penalty provides.

\begin{figure}[!b]
    \begin{center}
        \includegraphics[width=1.0\textwidth]{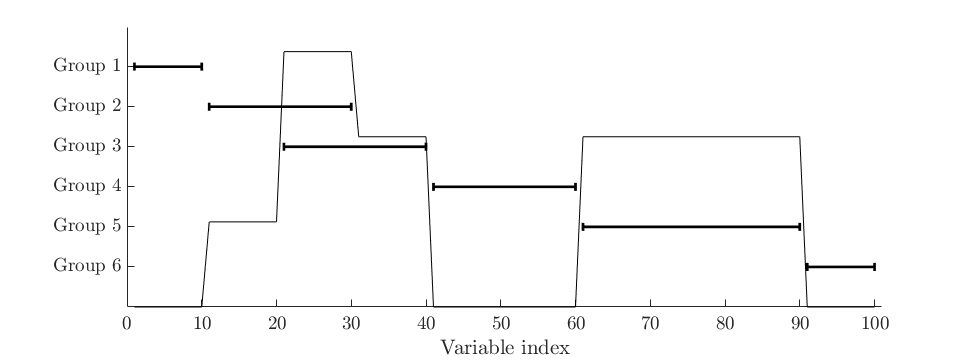}
        \caption{An illustration of the variable groups in $\X_2$ for the simulated
                 data example. The horizontal thick lines correspond to a group, and
                 the variables included in a group are indicated on the first axis.
                 The underlying, true, weight vectors is superimposed with thin
                 lines.}
        \label{fig:gl_groups}
    \end{center}
\end{figure}

The groups of the group \(\ell_{1,2}\) penalty were: \(\mathrm{Group~1}=[1,\ldots,10]\),
\(\mathrm{Group~2}\) \(=[11,\ldots,30]\), \(\mathrm{Group~3}=[21,\ldots,40]\),
\(\mathrm{Group~4}=[41,\ldots,60]\), \(\mathrm{Group~5}=[61,\ldots,90]\)
and \(\mathrm{Group~6}=[91,\ldots,100]\), as illustrated in \figref{fig:gl_groups}.
Note in particular that Group 2 and Group 3 are overlapping.

The regularisation constants, \(\tau_1\) and \(\tau_2\), were computed using the
method of Sch\"{a}fer and Strimmer~\cite{Schafer_and_Strimmer_2005}, and were
\(\tau_1=0.33\) and \(\tau_3=0.32\), for the blocks \(\X_1\) and \(\X_2\),
respectively.

We performed a grid search with cross-validation, as in the previous example, in
order to find the regularisation parameters. The constants for the regularised RGCCA
model were found to be \(\omega_1=0.61\), \(s_1=7.7\) and \(\omega_2=0.13\).

The optimisation problem that we considered in this example was therefore
\begin{align}
    \minimise_{\substack{\w_k\in\mathbb{R}^{p_k},\\k=1,2}} \;& \widehat{f}(\v{w}_1,\v{w}_2) = -\text{Cov}(\X_1\v{w}_1, \X_2\v{w}_2) \nonumber\\[-1em]
                                      &\qquad\qquad\quad\;\, + 0.61\cdot\widehat{\Omega}_{TV}(\mu_1, \v{w}_1) + 0.13\cdot\widehat{\Omega}_{GL}(\mu_2, \v{w}_2), \nonumber\\
    \mathrm{subject~to} \;& \w_1 \in \{\x \in \mathbb{R}^{p_1} \;|\; 0.33\cdot\|\x\|_2^2 + 0.67\cdot\text{Var}(\X_2\x) \leq 1 \land \|\x\|_1 \leq 7.7\}, \nonumber\\
    \phantom{subject~to} \;& \w_2 \in \{\x \in \mathbb{R}^{p_2} \;|\; 0.32\cdot\|\x\|_2^2 + 0.68\cdot\text{Var}(\X_2\x) \leq 1\}, \nonumber
\end{align}
in which \(\mu_1=\mu_2=5\cdot10^{-4}\). We only extracted one component in this example.

\section{Results}

\subsection{Glioma data set}

The locations were predicted using three approach\-es: From the GE data, \(\X_1\),
from the CGH data, \(\X_2\), and from both the GE and the CGH data concatenated,
\ie from \(\X_{12} = [\X_1 \,|\, \X_2]\). The GE data, \(\X_1\), were able to predict
\(43 / 53 \approx 81~\%\) of the locations correctly; the CGH data, \(\X_2\), were
able to predict \(38 / 53 \approx 72~\%\) of the locations correctly; and when
simultaneously predicting from both the GE and the CGH data, \(50 / 53 \approx 94~\%\)
of the locations were correctly identified. These numbers are the means of the
7-folds of cross-validated prediction rates. These prediction rates are similar to or
higher than those reported in~\cite{Tenenhaus_etal_2014} and in the present case,
structure was also imposed on the weight vectors.


We performed 100 bootstrap rounds in order to assess how stable the models were, and
in particular whether the models are in agreement or not between bootstrap rounds.

We computed an inter-rater agreement measure: the Fleiss' \(\kappa\)
statistic~\cite{Fleiss_1971}, in order to asses the agreement between weight
vectors computed from the 100 bootstrap rounds in terms of the selected variables
for the two components. The number of times a variable is selected or not selected
in the 100 bootstrap rounds is counted and summarised by the Fleiss' \(\kappa\),
that thus measures the agreement among the bootstrap samples. The higher the value
of \(\kappa\), the more stable the method is with respect to sampling; positive
values means more stable than what could be expected from chance, and \(\kappa=1\)
means completely stable (all samples agree entirely).

The models were stable, with all Fleiss' \(\kappa>0\). \Ie, the weight vectors
computed in the different bootstrap rounds agreed in terms of which variables should
be included or not in the model. Fleiss' \(\kappa\) was about \(0.61\) for the first
component of \(\X_1\), and \(0.35\) for the second component; about \(0.28\) for the
first component of \(\X_2\) and \(0.07\) for the second component.

\begin{figure}[!b]
    \begin{center}
        \includegraphics[width=0.49\textwidth]{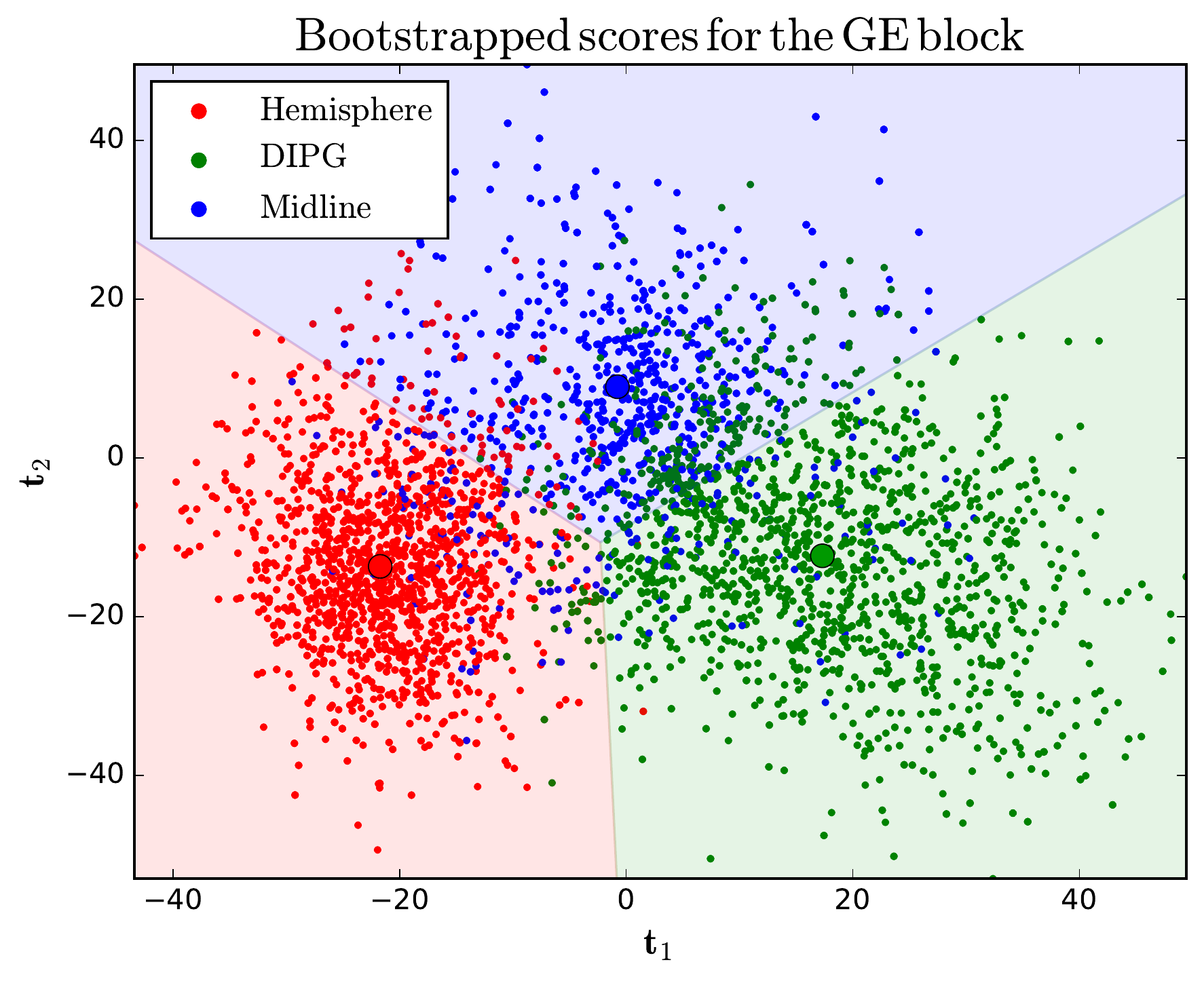}%
        \includegraphics[width=0.49\textwidth]{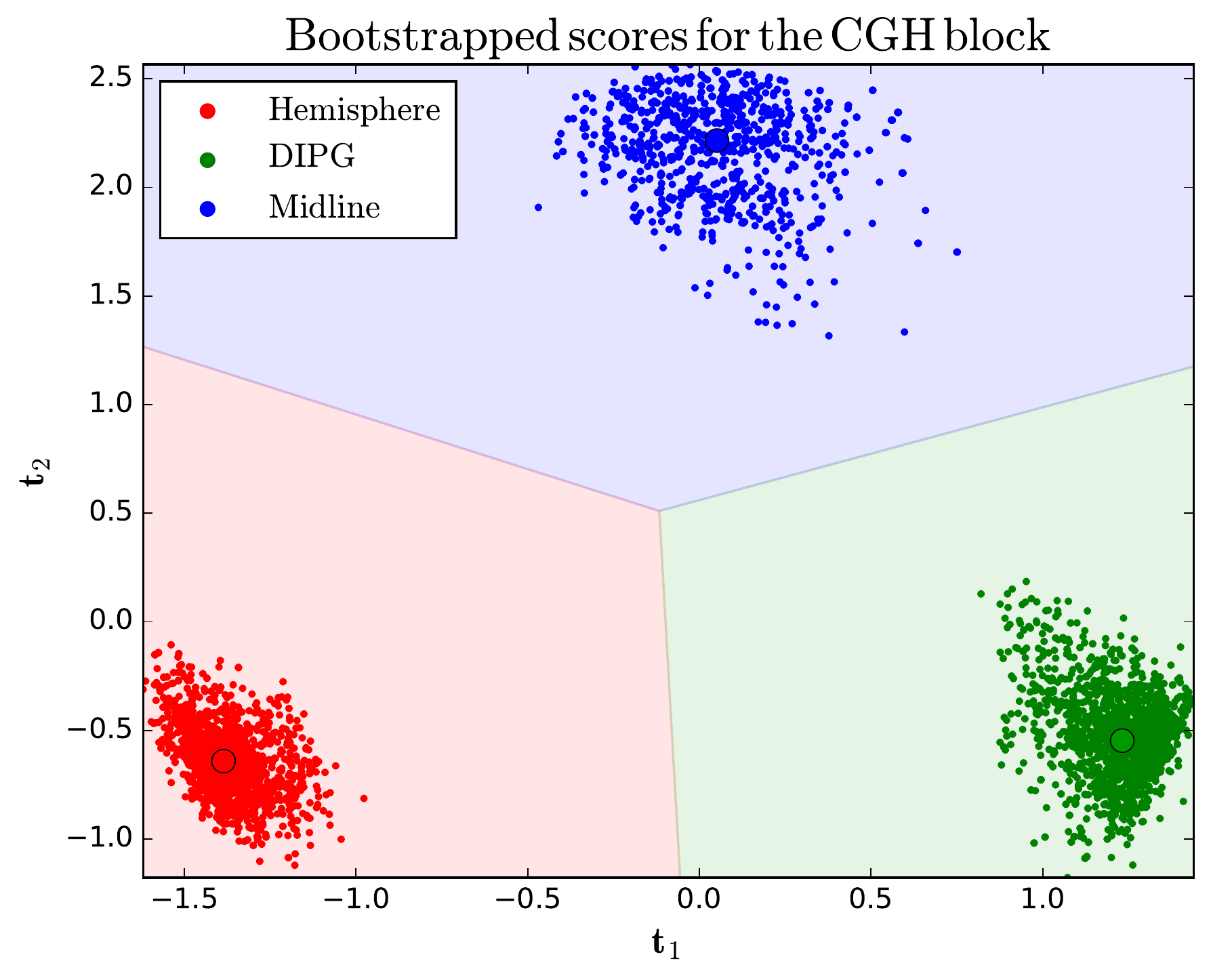}
        \caption{The first and second score vectors of the predictions of the
                 samples from the 100 bootstrap rounds for the GE data (left) and
                 the CGH data (right). The bootstrapped means are indicated and the
                 Voronoi regions illustrate the separation.}
        \label{fig:bootstrapped_score_plot}
    \end{center}
\end{figure}

\figref{fig:bootstrapped_score_plot} illustrates the first and second components of
the models from the samples of the 100 bootstrap rounds; it is clear that the CGH
data discriminates the locations very well. The coloured sections are Voronoi regions,
and are meant to illustrate the location separation.

\begin{figure}[!b]
    \begin{center}
        \includegraphics[width=0.49\textwidth]{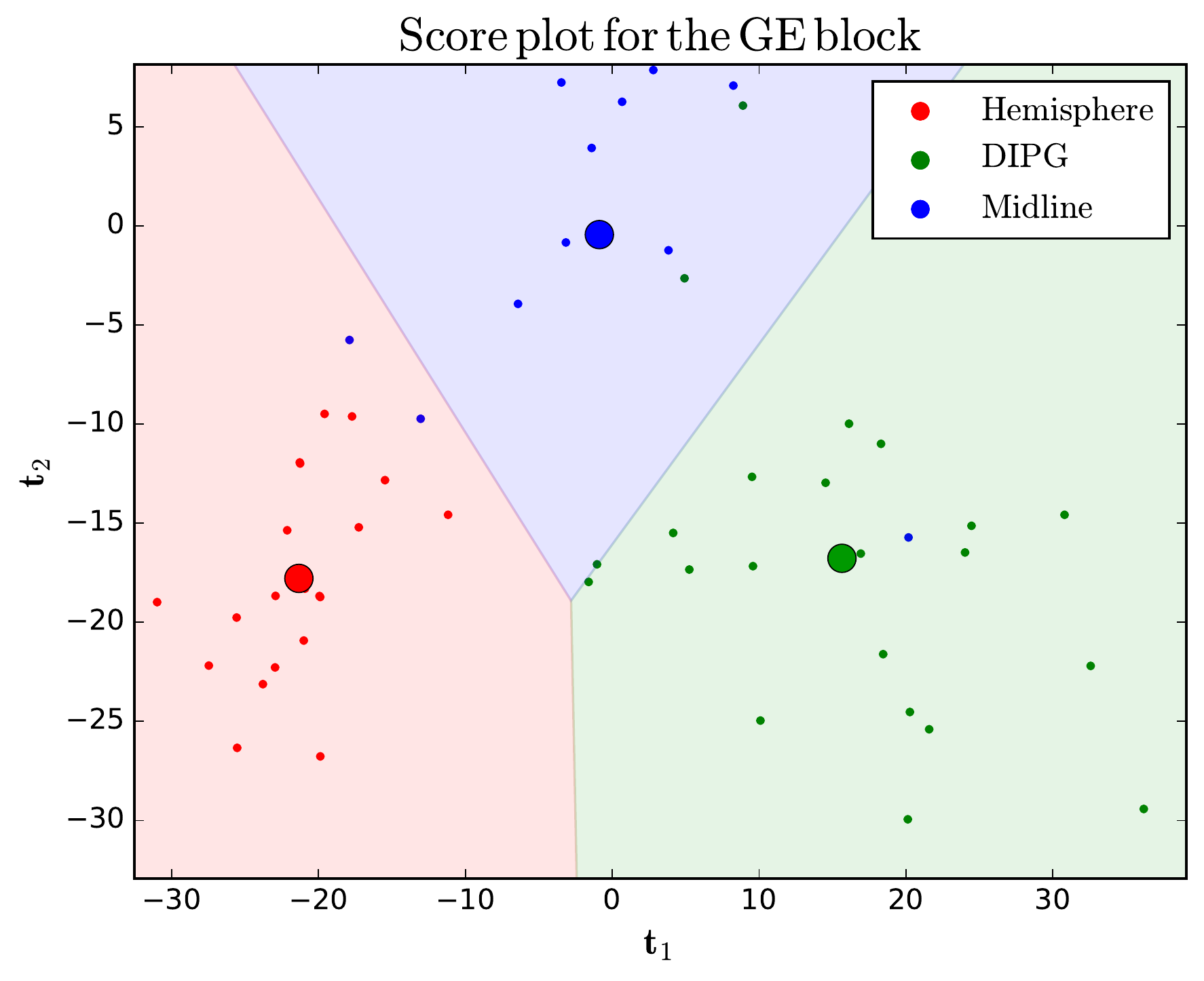}%
        \includegraphics[width=0.49\textwidth]{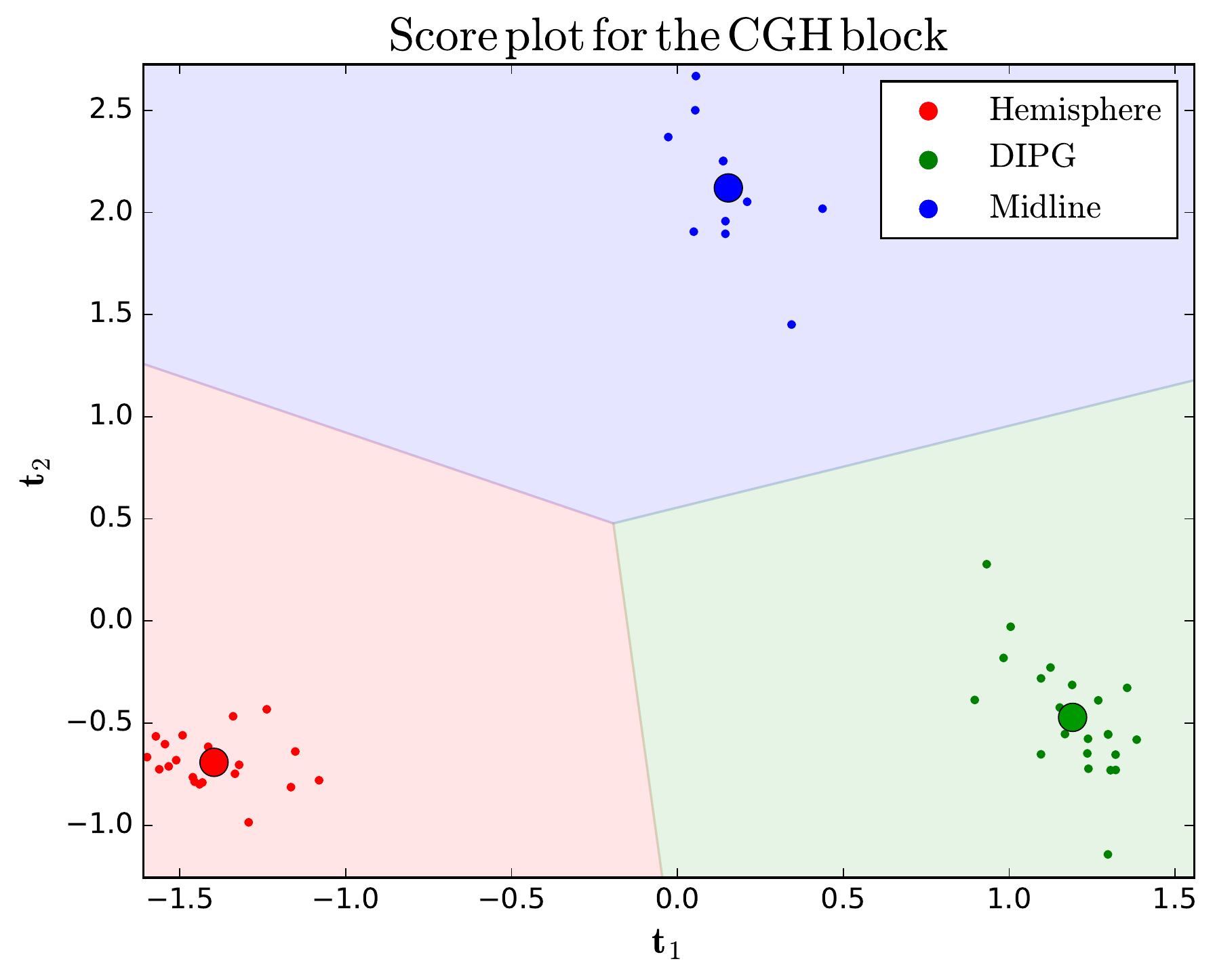}\\
        \includegraphics[width=0.49\textwidth]{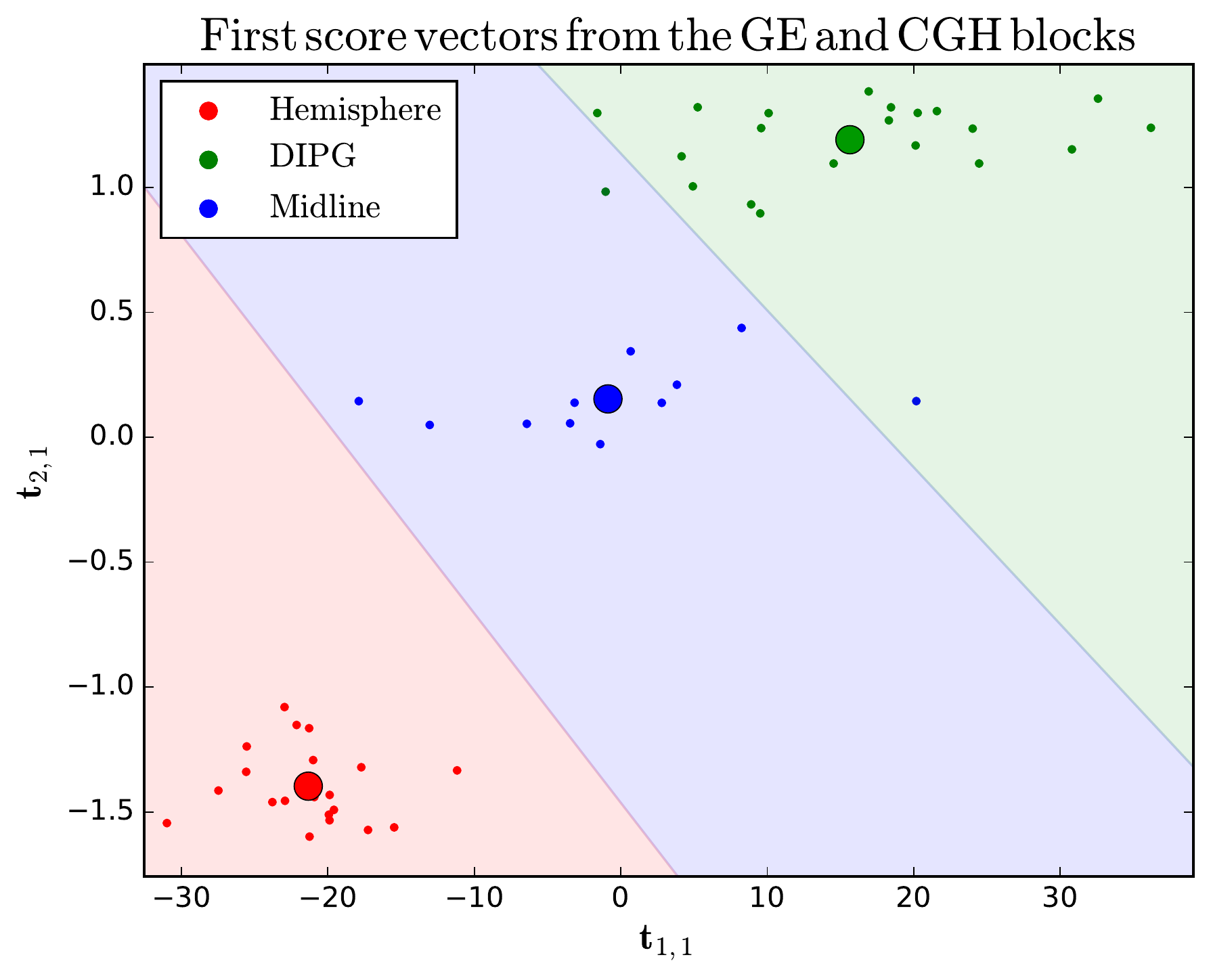}
        \caption{The first and second score vectors of the final penalised RGCCA
                 model for the GE data (left) and the CGH data (right). The bottom
                 plot illustrates the first score vectors plotted against each other.
                 The predicted means are indicated in the plot by larger circles.}
        \label{fig:score_plot}
    \end{center}
\end{figure}

\figref{fig:score_plot} illustrates the first and second component of the
final RGCCA model for the GE (upper left panel) and CGH (upper right panel) blocks.
The lower centre panel illustrates the first components of the GE and CGH blocks
plotted against each other. The separation between the locations is clear. It is
also clear that the score vectors of the GE and CGH data are correlated, and that
the variance of the model of the GE data is higher.

The bootstrap average number of selected variables in \(\X_1\) was roughly
3~\% in both components. In \(\X_2\), the average number of selected variables
was roughly 27~\% in the first component, and roughly 40~\% in the second
component.

The group \(\ell_{1,2}\) penalty selected \(125.5\) out of the 199 identified
groups in the first component and \(126.3\) in the second component (these values
are bootstrap averages). Groups were considered \textit{strong} if they had a high
ratio between the number of selected gene expressions within the group over the
total number of selected gene expressions~\cite{Chen_etal_2012}.

Among the top ranking groups were:
Alzheimer's disease (hsa05010), which implies a relation to a supratentorial
tumour (in the hemispheres) since it affects the cortex and hypocampus;
``Axon guidance'' (hsa04360), which implies a relation to DIPG, because of the
abundance of axons in the brain stem; and
Nucleotide excision repair (hsa03420), since it could explain tendencies towards
drug resistance.

Among the groups that were excluded from the model were the
Citrate cycle (TCA cycle, hsa00020). Citrate seems to be abundant in DIPG
(unpublished results), but its occurrence in other locations is unknown. This would
imply that it could be similarly found in the other locations or cancer types as
well, and thus be a poor predictor of the tumour locations.

\subsection{Simulated data}

\begin{figure}[!bh]
    \begin{center}
        \includegraphics[width=1.0\textwidth]{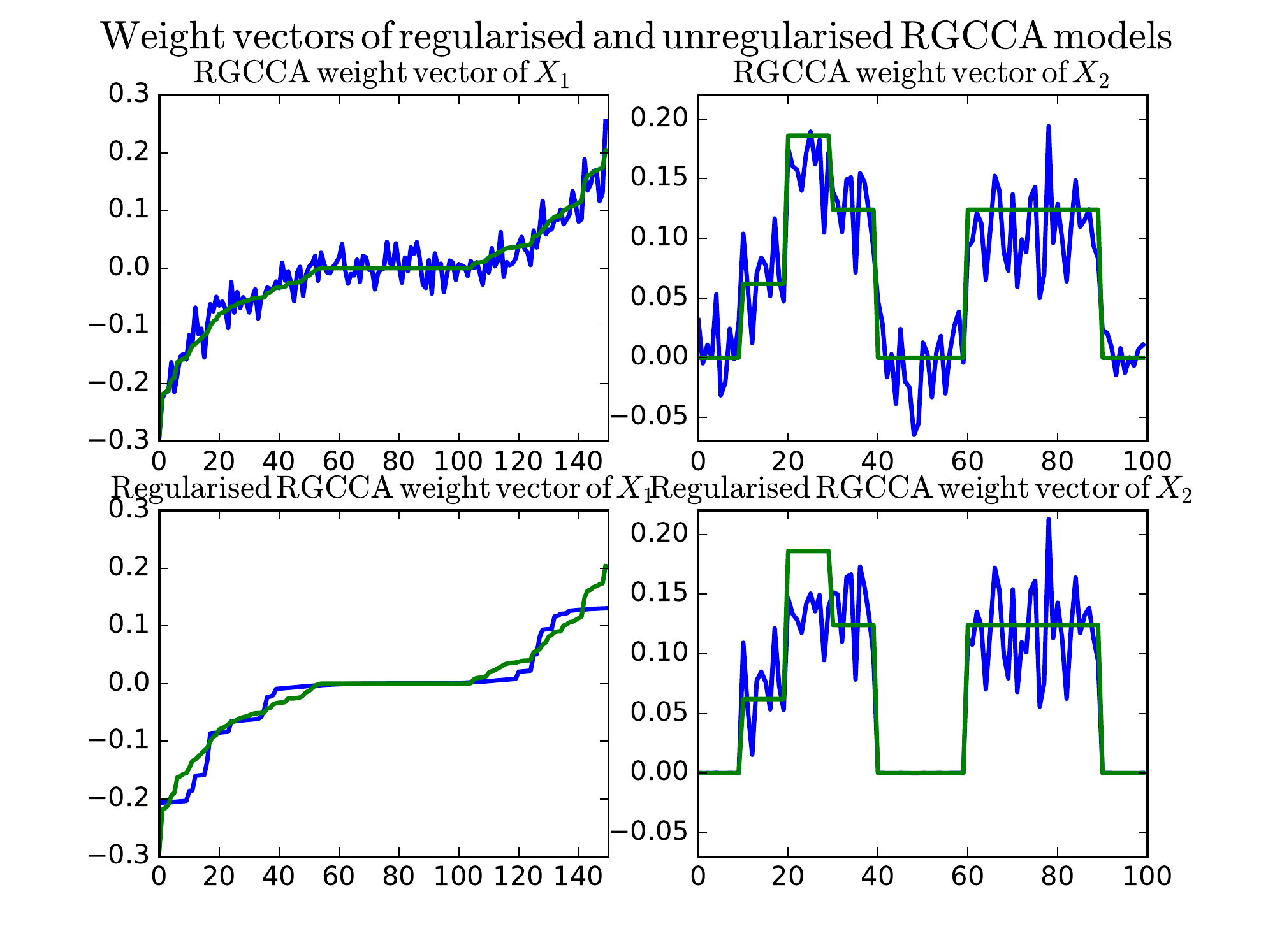}
        \caption{The weight vectors for an unpenalised RGCCA model (top row) for the two
                 blocks and the weight vectors for an RGCCA model with total variation,
                 $\ell_1$ and group lasso penalties (bottom row). The total variation
                 penalty clearly performs well in reconstructing the true underlying weight
                 vectors (plotted in green). The group lasso penalty effectively cancels the
                 groups with true null weights, and appears to capture the true group means
                 fairly well, but is affected by noise. Note also that the variables with
                 index 21 through 30 in $\X_2$ belongs to two groups.}
        \label{fig:sim_results}
    \end{center}
\end{figure}

We performed cross-validation in order to find the parameters for the total variation,
$\ell_1$ and group lasso penalties. The obtained weight vectors were compared to the
true ones (those used to generate the data) and to those of an unpenalised, or
``regular'', RGCCA model. The weight vectors are illustrated in
\figref{fig:sim_results}.

The weights from the unpenalised RGCCA model nicely capture the trends in the data, but are
affected to a high degree by the noise.

As seen in \figref{fig:sim_results}, the weight vectors obtained from the penalised model
clearly performs well in reconstructing the true weights; especially so for $\X_1$, where
the total variation penalty removes much of the noise, but leaves the true weight profile
intact. The group lasso penalty manages to remove the groups with true null weights, and
appears to find the mean of the other groups.

The weights of $\X_2$ that correspond to variable indices 20 through 30 belong to two
groups (group 2 and group 3). Those should have weights that are a compromise between
describing group 2 and describing group 3. It appears they focus on describing group 2.

\section{Discussion and conclusions}

The proposed method solves a restriction in the SGCCA method that prevents the
regularisation parameters, \(\tau_k\), for \(k=1,\ldots,K\), to take other values
than one. The proposed method doesn't have this restriction, and allows the
\(\tau_k\) to vary between zero and one. We also showed how to extend the RGCCA
and SGCCA methods by allowing complex structured penalties to be included in the
model.

The proposed generalised RGCCA method was applied to gene expression and
CGH data to predict the location of tumours in glioma. We used a group
\(\ell_{1,2}\) penalty on the GE data and a total variation penalty on the CGH
data. Both data sets were also subject to an \(\ell_1\) constraint and the
quadratic constraint used in RGCCA (a generalised norm constraint). The results are
very encouraging and illustrate the importance of structured constraints.

The proposed method was also applied to simulated data were it was shown that it
performed well in reconstruct the \textit{true} weight vectors that were used in
constructing the simulated data.

The authors intend to resolve the restriction that $g(x) = x$ (Horst's scheme) in
\eqref{eq:loss_function_structure} in future research in order to be able to
formulate even more general models. Future work also include adapting the CONESTA
algorithm~\cite{Hadj-Selem_etal_2016_preprint} to the present problem formulation,
in order to obtain results faster, and with higher precision.

The proposed minimisation problem comprise many well-known multiblock and PLS-based
methods as special cases. Examples include PLS-R, Sparse PLS, PCA, Sparse PCA, CCA,
RGCCA, SGCCA, \textit{etc.}, but the proposed method has the advantage that it
allows structured and sparsity-inducing penalties to be included in the model.


\section{Acknowledgements}

This work was supported by grants from the French National Research Agency: ANR
IA BRAINOMICS (ANR-10-BINF-04), and a European Commission grant: MESCOG (FP6
ERA-NET NEURON 01 EW1207).

\bibliographystyle{plain}
\bibliography{multistruct}

\begin{thebibliography}{10}

\bibitem{Beck_Teboulle_2009}
Amir Beck and Marc Teboulle.
\newblock {A Fast Iterative Shrinkage-Thresholding Algorithm for Linear Inverse
  Problems}.
\newblock {\em SIAM Journal on Imaging Sciences}, 2(1):183--202, Jan 2009.

\bibitem{Beck_Teboulle_2009b}
Amir Beck and Marc Teboulle.
\newblock Gradient-based algorithms with applications to signal recovery
  problems.
\newblock In Daniel~P. Palomar and Yonina~C. Eldar, editors, {\em Convex
  Optimization in Signal Processing and Communications}, chapter~2, pages
  42--88. Cambridge University Press, 1 edition, 2009.

\bibitem{Chen_2011}
Xi~Chen and Han Liu.
\newblock {An Efficient Optimization Algorithm for Structured Sparse CCA, with
  Applications to eQTL Mapping}.
\newblock {\em Statistics in Biosciences}, 4(1):3--26, December 2011.

\bibitem{Chen_etal_2012}
Xi~Chen, Han Liu, and Jaime~G. Carbonell.
\newblock Structured sparse canonical correlation analysis.
\newblock volume~22 of {\em JMLR Workshop and Conference Proceedings}, pages
  199--207, April 2012.
\newblock AISTATS 2012.

\bibitem{Combettes_Pesquet_2011}
Patrick~L. Combettes and Jean-Christophe Pesquet.
\newblock Proximal splitting methods in signal processing.
\newblock In H.~H. Bauschke, R.~S. Burachik, P.~L. Combettes, V.~Elser, D.~R.
  Luke, and H.~Wolkowicz, editors, {\em Fixed-Point Algorithms for Inverse
  Problems in Science and Engineering}, pages 185--212. New York: Springer,
  2011.

\bibitem{DeLeeuw_1994}
Jan De~Leeuw.
\newblock Block relaxation algorithms in statistics.
\newblock In H.~H. Bock, W.~Lenski, and M.~M. Richter, editors, {\em
  Information Systems and Data Analysis}, pages 308--325. Springer, Berlin,
  1994.

\bibitem{Fleiss_1971}
Joseph~L. Fleiss.
\newblock Measuring nominal scale agreement among many raters.
\newblock {\em Psychological Bulletin}, 76(5):378--382, 1971.

\bibitem{Hadj-Selem_etal_2016_preprint}
Fouad Hadj-Selem, Tommy L\"ofstedt, Vincent Frouin, Vincent Guillemot, and
  Edouard Duchesnay.
\newblock {An Iterative Smoothing Algorithm for Regression with Structured
  Sparsity}.
\newblock {\em arXiv:1605.09658 [stat]}, May 2016.

\bibitem{Lofstedt_etal_2016}
Tommy L\"{o}fstedt, Fouad Hadj-Selem, Vincent Guillemot, Cathy Philippe,
  Edouard Duchesnay, Vincent Frouin, and Arthur Tenenhaus.
\newblock Structured variable selection for regularized generalized canonical
  correlation analysis.
\newblock In Herv\'e Abdi, Vincenzo Esposito~Vinzi, Giorgio Russolillo, Gilbert
  Saporta, and Laura Trinchera, editors, {\em The Multiple Facets of Partial
  Least Squares Methods}, volume 173 of {\em Springer Proceedings in
  Mathematics \& Statistics}. Springer, 2016.

\bibitem{Michel_etal_2011}
Vincent Michel, Alexandre Gramfort, Ga\"{e}l Varoquaux, Evelyn Eger, and
  Bertrand Thirion.
\newblock {Total Variation Regularization for fMRI-Based Prediction of
  Behavior}.
\newblock {\em IEEE Transactions on Medical Imaging}, 30(7):1328--1340, 2011.

\bibitem{Nesterov_2004}
Yurii Nesterov.
\newblock {Smooth minimization of non-smooth functions}.
\newblock {\em Mathematical Programming}, 103(1):127--152, December 2004.

\bibitem{Parikh_Boyd_2013}
Neal Parikh and Stephen Boyd.
\newblock {\em Proximal Algorithms}.
\newblock Foundations and Trends in Optimization. Now Publishers Inc., 1st
  edition, 2013.

\bibitem{Philippe_2012}
Cathy Philippe, Stephanie Puget, Dorine~A Bax, Bastien Job, Pascale Varlet,
  Marie-Pierre Junier, Felipe Andreiuolo, Dina Carvalho, Ricardo Reis, Lea
  Guerrini-Rousseau, Thomas Roujeau, Philippe Dessen, Catherine Richon,
  Vladimir Lazar, Gwenael {Le Teuff}, Christian Sainte-Rose, Birgit Geoerger,
  Gilles Vassal, Chris Jones, and Jacques Grill.
\newblock {Mesenchymal transition and PDGFRA amplification/mutation are key
  distinct oncogenic events in pediatric diffuse intrinsic pontine gliomas.}
\newblock {\em PloS one}, 7(2):e30313, January 2012.

\bibitem{Qin_etal_2013}
Zhiwei Qin, Katya Scheinberg, and Donald Goldfarb.
\newblock {Efficient block-coordinate descent algorithms for the Group Lasso}.
\newblock {\em Mathematical Programming Computation}, 5(2):143--169, Mar 2013.

\bibitem{Sanchez_2013}
Gaston Sanchez.
\newblock {PLS Path Modeling with R}.
\newblock \url{http://www.gastonsanchez.com/PLS_Path_Modeling_with_R.pdf},
  2013.

\bibitem{Schafer_and_Strimmer_2005}
Juliane Sch\"{a}fer and Korbinian Strimmer.
\newblock A shrinkage approach to large-scale covariance matrix estimation and
  implications for functional genomics.
\newblock {\em Statistical applications in genetics and molecular biology},
  4(1), 2005.

\bibitem{Schmidt_etal_2011}
Mark Schmidt, Nicolas Le~Roux, and Francis Bach.
\newblock Convergence rates of inexact proximal-gradient methods for convex
  optimization.
\newblock arXiv:1109.2415, September 2011.

\bibitem{Silver_and_Giovanni_2012}
Matt Silver and Giovanni Montana.
\newblock {Fast identification of biological pathways associated with a
  quantitative trait using group lasso with overlaps.}
\newblock {\em Statistical applications in genetics and molecular biology},
  11(1):Article 7, January 2012.

\bibitem{Tenenhaus_etal_2014}
Arthur Tenenhaus, Cathy Philippe, Vincent Guillemot, Kim-Anh L\^e~Cao, Jacques
  Grill, and Vincent Frouin.
\newblock {Variable Selection For Generalized Canonical Correlation Analysis}.
\newblock 2014.

\bibitem{Tenenhaus_and_Tenenhaus_2011}
Arthur Tenenhaus and Michel Tenenhaus.
\newblock {Regularized Generalized Canonical Correlation Analysis}.
\newblock {\em Psychometrika}, 76(2):257--284, 2011.

\bibitem{Tenenhaus_etal_2005}
Michel Tenenhaus, Vincenzo Esposito~Vinzi, Yves-Marie Chatelin, and Carlo
  Lauro.
\newblock {PLS path modeling}.
\newblock {\em Computational Statistics \& Data Analysis}, 48:159--205, 2005.

\bibitem{van_den_Berg_etal_2008}
Ewout van~den Berg, Mark Schmidt, Michael~P. Friedlander, and Kevin Murphy.
\newblock Group sparsity via linear-time projection.
\newblock Technical Report TR-2008-09, Department of Computer Science,
  University of British Columbia, Vancouver, Canada, June 2008.

\bibitem{Vinod_1976}
Hrishikesh~D. Vinod.
\newblock {Canonical ridge and econometrics of joint production}.
\newblock {\em Journal of Econometrics}, 4:147--166, 1976.

\bibitem{Wegelin_2000}
Jacob~A. Wegelin.
\newblock A survey of partial least squares (pls) methods, with emphasis on the
  two-block case.
\newblock Technical Report 371, Department of Statistics, University of
  Washington, Seattle, Washington, USA., March 2000.

\bibitem{Witten_etal_2009}
Daniela~M. Witten, Robert Tibshirani, and Trevor Hastie.
\newblock A penalized matrix decomposition, with applications to sparse
  principal components and canonical correlation analysis.
\newblock {\em Biostatistics}, 10(3):515--534, 2009.

\end{thebibliography}

\appendix

\section{The RGCCA constraint} \label{apx:RGCCA_constraint}

Let \(\x\) be a \(p \times 1\) real vector and let \(\M{M}\) be a symmetric positive
(possibly semi-) definite matrix. We consider the following optimisation problem
\begin{align} \label{apx:eq:generalised_norm}
    \minimise_{\y \in \mathbb{R}^p} \;\;& \frac{1}{2}\|\y - \x\|^2_2 \\
    \text{subject to}               \;\;& \y^\T\M{M}\y \leq c. \nonumber
\end{align}
This problem is equivalent to a projection of the point \(\x\) onto a hyperellipse (a multi-dimensional
ellipse) whose equation is \(\y^\T\M{M}\y = c\), \ie a hyperellipse defined by \(\M{M}\) with
radius \(c\). We will thus assume that \(\x^\T\M{M}\x > c\), since otherwise the
problem is trivial and the solution is \(\x\). This assumption also implies that
there is a single unique solution to this problem.

Let \(\X\) be an \(n \times p\) real matrix and \(\sigma_1, \sigma_2, \ldots, \sigma_{\min(n, p)}\) be its
singular values, then
\[
    \lambda_i = \begin{cases}
                    \frac{1-\tau}{n-1}\sigma^2_i + \tau & \text{if } i \leq \min(n, p), \\
                    \tau                                & \text{if } \min(n, p) < i \leq p,
                \end{cases}
\]
for \(i=1,\ldots,p\), are the eigenvalues of \(\M{M} = \tau \mathbf{I}_p + \frac{(1-\tau)}{n-1} \X^\T \X\). Since \(\M{M}\) is a real symmetric matrix, there exists an orthogonal
matrix \(\M{P}\) such that $\M{P}^{-1}\M{M}\M{P} = \M{\Lambda}$, where \(\M{\Lambda}\) is a diagonal
matrix that contains the eigenvalues of \(\M{M}\).

We now define $\widetilde{\x} = \M{P}^{-1}\x$ and $\widetilde{\y} = \M{P}^{-1}\y$ and can
thus solve a much simpler problem
\begin{align}
    \minimise_{\widetilde{\y} \in \mathbb{R}^p} \;\;& \frac{1}{2}\|\widetilde{\y} - \widetilde{\x}\|^2_2 \nonumber \\
    \text{subject to}                           \;\;& \widetilde{\y}^\T\M{\Lambda}\widetilde{\y} \leq c. \nonumber
\end{align}

The Lagrange formulation of this optimisation problem is
\[
    \mathcal{L}(\widetilde{\y}, \gamma) = \frac{1}{2}(\widetilde{\y} - \widetilde{\x})^\T(\widetilde{\y} - \widetilde{\x})
        + \gamma(\widetilde{\y}^\T\M{\Lambda}\widetilde{\y} - c).
\]
Cancelling the gradient of the Lagrangian function $\mathcal{L}$ with respect to $\widetilde{\y}$
yields the following stationary equations
\[
    \widetilde{\y} - \widetilde{\x} + 2\gamma\M{\Lambda}\widetilde{\y} = \v{0},
\]
and the solution is obtained as
\[
    \widetilde{\y} = \left(\I_p + 2\gamma\M{\Lambda}\right)^{-1}\widetilde{\x}
                   = \diag\left(\frac{1}{1 + 2\gamma\lambda_1}, \ldots, \frac{1}{1 + 2\gamma\lambda_p}\right)\widetilde{\x},
\]
for some \(\gamma\).

Thus,
\[
    \widetilde{\y}^\T\M{\Lambda}\widetilde{\y} = c
\]
if and only if
\[
    \widetilde{\x}^\T\diag\left(\frac{\lambda_1}{(1 + 2\gamma\lambda_1)^2}, \ldots, \frac{\lambda_p}{(1 + 2\gamma\lambda_p)^2}\right)\widetilde{\x} = c.
\]

Hence, we form the auxiliary function
\[
    f(\gamma) = \sum_{i=1}^p \widetilde{x}_i^2 \frac{\lambda_i}{(1 + 2\gamma\lambda_i)^2} - c.
\]
with derivative
\[
    f'(\gamma) = -4\sum_{i=1}^p \widetilde{x}_i^2 \frac{\lambda_i^2}{(1 + 2\gamma\lambda_i)^3}.
\]

However, we note that this approach requires us to compute all the \(p\)
eigenvectors of \(\M{P}\), in order to find all \(\widetilde{x}_i\), which may
be computationally infeasible. We use the following trick to go around this
problem: We note that the \(p - \min(n, p)\) eigenvalues are all equal to
\(\tau\), which means that
\[
    \sum_{i > \min(n, p)} \widetilde{x}_i^2 \frac{\lambda_i}{(1 + 2\gamma\lambda_i)^2}
        = \frac{\tau}{(1 + 2\gamma\tau)^2} \sum_{i > \min(n, p)} \widetilde{x}_i^2.
\]
Moreover, since \(\M{P}\) is an orthogonal matrix,
\[
    \sum_{i > \min(n, p)} \widetilde{x}_i^2
        = \|\x\|_2^2 - \sum_{i \leq \min(n, p)} \widetilde{x}_i^2,
\]
and hence we implicitly know the values of \(\widetilde{x}_i\) for
\(i > \min(n, p)\), without computing and multiplying by the corresponding
eigenvectors.

We may thus rewrite the auxiliary function as
\begin{align}
    f(\gamma) &= \sum_{i=1}^p \widetilde{x}_i^2 \frac{\lambda_i}{(1 + 2\gamma\lambda_i)^2} - c \nonumber\\
              &= \frac{\tau}{(1 + 2\gamma\tau)^2} \left[\|\x\|_2^2 - \sum_{i \leq \min(n, p)} \widetilde{x}_i^2\right]
                + \sum_{i \leq \min(n, p)} \widetilde{x}_i^2 \frac{\lambda_i}{(1 + 2\gamma\lambda_i)^2}
                - c, \nonumber
\end{align}
with derivative
\[
    f'(\gamma) = -4\frac{\tau^2}{(1 + 2\gamma\tau)^3} \left[\|\x\|_2^2 - \sum_{i \leq \min(n, p)} \widetilde{x}_i^2\right]
                - 4\sum_{i \leq \min(n, p)} \widetilde{x}_i^2 \frac{\lambda_i^2}{(1 + 2\gamma\lambda_i)^3}.
\]

Now \(\gamma\) can be found numerically by the Newton-Raphson method. We set an
initial value of \(\gamma^{(0)}\), \eg \(\gamma^{(0)} = 0\), and compute the
sequence \((\gamma^{(s)})_s\) iteratively by
\[
    \gamma^{(s+1)} = \gamma^{(s)} - \frac{f(\gamma^{(s)})}{f'(\gamma^{(s)})}.
\]
The algorithm stops when \(|\gamma^{(s+1)} - \gamma^{(s)}| < \varepsilon\),
for some small \(\varepsilon\), \eg \(\varepsilon = 5\cdot10^{-16}\). We denote the
final element of the sequence by \(\gamma^*\).

We recall that the proximal operator of the quadratic function in
\eqref{apx:eq:generalised_norm} is
\begin{align}
    \prox_{\gamma}(\x) &= \argmin_{\y\in\mathbb{R}^p} \frac{1}{2}\|\y - \x\|_2^2 + \gamma(\y^\T\M{M}\y - c) \nonumber\\
                       &= \left(\I_p + 2\gamma\M{M}\right)^{-1}\x. \nonumber
\end{align}
Thus, we have the projection
\[
    \proj_{\mathcal{S}}(\x) = \prox_{\gamma^*}(\x),
\]
where \(\mathcal{S} = \{\x \in \mathbb{R}^p \;|\; \x^\T\M{M}\x \leq c\}\).
Hence, the projection is computed as
\[
    \y = \left(\I_p + 2\gamma^*\M{M}\right)^{-1}\x,
\]
where the inverse is computed only once, and can be computed efficiently by using the
Woodbury matrix identity.

\end{document}